\documentclass{article} 
\usepackage{iclr2019_conference,times}

\usepackage{amsmath,amsfonts,bm}

\def\eqref#1{equation~\ref{#1}}

\def\1{\bm{1}}

\DeclareMathAlphabet{\mathsfit}{\encodingdefault}{\sfdefault}{m}{sl}
\SetMathAlphabet{\mathsfit}{bold}{\encodingdefault}{\sfdefault}{bx}{n}

\usepackage[T1]{fontenc}
\usepackage{subcaption}
\usepackage{graphicx}
\usepackage{hyperref}
\usepackage{csquotes}
\usepackage{url}

\title{Bridging Adversarial Robustness and \\ Gradient Interpretability}

\author{Beomsu Kim \thanks{This work was done while Beomsu Kim was an intern at SI Analytics.} \\
School of Computing \\
Korea Advanced Institute of Science and Technology \\
Daejeon, Republic of Korea \\
\texttt{3141kbs@kaist.ac.kr} \\
\And
Junghoon Seo \\
Satrec Initiative \\
Daejeon, Republic of Korea \\
\texttt{sjh@satreci.com} \\
\AND
Taegyun Jeon \\
SI Analytics \\
Daejeon, Republic of Korea \\
\texttt{tgjeon@si-analytics.ai}
}

\iclrfinalcopy 
\begin{document}

\maketitle

\begin{abstract}
Adversarial training is a training scheme designed to counter adversarial attacks by augmenting the training dataset with adversarial examples. Surprisingly, several studies have observed that loss gradients from adversarially trained DNNs are visually more interpretable than those from standard DNNs. Although this phenomenon is interesting, there are only few works that have offered an explanation. In this paper, we attempted to bridge this gap between adversarial robustness and gradient interpretability. To this end, we identified that loss gradients from adversarially trained DNNs align better with human perception because adversarial training restricts gradients closer to the image manifold. We then demonstrated that adversarial training causes loss gradients to be quantitatively meaningful. Finally, we showed that under the adversarial training framework, there exists an empirical trade-off between test accuracy and loss gradient interpretability and proposed two potential approaches to resolving this trade-off.
\end{abstract}

\section{Introduction} \label{section:introduction}

\textit{Adversarial attack} is an imperceptible perturbation to the input image which causes a deep neural network (DNN) to misclassify the perturbed input image with high confidence \citep{Goodfellow2015}. Such perturbed inputs are called \textit{adversarial examples}. Numerous defence approaches have been proposed to create \textit{adversarially robust} DNNs that are resistant to adversarial attacks. One of the most common and successful defence methods is \textit{adversarial training}, which augments the training dataset with adversarial examples \citep{Szegedy2013}.

Surprisingly, several studies have observed that loss gradients from adversarially trained DNNs are \textit{visually} more interpretable than those from standard DNNs, i.e., DNN trained on natural images \citep{Ross2018,Tsipras2018,Zhang2019}. These studies have used the visual interpretability of gradients as evidence that adversarial training causes DNNs to learn meaningful feature representations which align well with salient data characteristics or human perception.

However, asking whether gradient visualization is meaningful to a human may be entirely different from determining whether it is an accurate description of the internal representation of the DNN. For instance, consider the DNN decision interpretation methods Deconvolution \citep{Zeiler2014} and Guided Backpropagation \citep{Springenberg2014}. They visualize the importance of each input pixel in the DNN decision process with heat maps produced by calculating imputed versions of the gradient. Although they produce sharp visualizations, they have been proven to be doing partial image recovery which is unrelated to DNN decisions \citep{Nie2018}.

In light of such studies, judging the interpretability of a DNN with the visual quality of its loss gradient runs the risk of being incorrect. There have been few attempts to analyze adversarial training in the context of DNN interpretability. For example, \citet{Chalasani2018} have investigated the effect of adversarial training on network weights. However, to the best of our knowledge, there is no work which provides thorough analyses of the effect of adversarial training on the visual and quantitative interpretability of DNN loss gradients.

In this paper, we attempted to bridge this gap between adversarial robustness and gradient interpretability through a series of experiments that addresses the following questions:\footnote{All codes for the experiments in this paper are public on \\ \url{https://github.com/1202kbs/Robustness-and-Interpretability}.} \enquote{why do loss gradients from adversarially trained networks align better with human perception?}, \enquote{is there a relation between the strength of adversary used for training and the perceptual quality of gradients?} and most importantly, \enquote{does adversarial training really cause loss gradients to better reflect the internal representation of the DNN?}. We then ended the paper by identifying a trade-off between test accuracy and gradient interpretability and proposing two potential approaches to resolving this trade-off. Specifically, we have the following three contributions:

\begin{enumerate}
\item \textbf{Visual Interpretability.} We showed that loss gradients from adversarially trained networks align better with human perception because adversarial training causes loss gradients to lie closer to the image manifold. We also provided a conjecture for this phenomenon and showed its plausibility with a toy dataset.
\item \textbf{Quantitative Interpretability.} We showed that loss gradients from adversarially trained networks are quantitatively meaningful. To this end, we established a formal framework for quantifying how accurately gradients reflect the internal representation of the DNN. We then verified whether gradients from adversarially trained networks are quantitatively meaningful using this framework.
\item \textbf{Accuracy and Interpretability Trade-off.} We showed with CNNs trained on CIFAR-10 that under the adversarial training framework, there exists an empirical trade-off between test accuracy and loss gradient interpretability. Then based on the experiment results, we proposed two potential approaches to resolving this trade-off.
\end{enumerate}

\section{Visual Interpretability of Loss Gradients} \label{section:visual}

We start by answering why adversarially trained DNNs have loss gradients that align better with human perception. All experiment settings in this paper can be found in the Appendix.

\subsection{Adversarial Training Confines Gradient to Image Manifold} \label{section:visual sub1}

To identify why adversarial training enhances the perceptual quality of loss gradients, we hypothesized adversarial examples from adversarially trained networks lie closer to the image manifold. Note that the loss gradient is the difference between the original image and the adversarial image. Hence if the adversarial image lies closer to the image manifold, the loss gradient should align better with human perception.

Following previous works \citet{Tsipras2018} and \citet{Stutz2018}, we first trained DNNs against adversarial attacks which maximize the training loss / cross entropy (XEnt) loss or the CW surrogate objective proposed by \citet{Carlini2016}. These objectives are maximized using Projected Gradient Descent (PGD) \citep{Kurakin2016} such that the adversarial examples stay within $\ell_2$ or $\ell_\infty$-distance of $\epsilon$ from the original image. We say an adversary, or an adversarial attack is stronger if its $\epsilon$ is larger. For consistency of observations, we trained the DNNs on three datasets: MNIST \citep{mnist1998}, FMNIST \citep{fmnist2017} and CIFAR-10 \citep{cifar2009}. Specific adversarial training procedures are described in Appendix \ref{section:2.1 settings}.

Then, we trained a VAE-GAN \citep{Larsen2016} on each dataset. Using its encoder $\mathrm{enc}$ and decoder $\mathrm{dec}$, we projected an image or an adversarial example $\mathbf{x}$ to the approximated manifold by $\pi(\mathbf{x}) = \mathrm{dec}(\mathrm{enc}(\mathbf{x}))$.\footnote{ We can also define the projection as $\pi(\mathbf{x}) = \mathrm{dec}(\mathbf{z}^*)$ where $\mathbf{z}^* = \arg\min_{\mathbf{z}} \|\mathbf{x} - \mathrm{dec}(\mathbf{z})\|_2$, but either definition led to the same results.} Next, we computed $d_\pi(\mathbf{x}) = \|\mathbf{x} - \pi(\mathbf{x})\|_2$ to quantify how close $\mathbf{x}$ is to the image manifold. Note that this concept of using a generative model to obtain an image's projection to the manifold has also been applied frequently in the context of adversarial defense \citep{Song2017,Meng2017,Samangoeui2018}.

\begin{figure}[t]
	\centering
	\begin{subfigure}{\linewidth}
	\includegraphics[width=\linewidth]{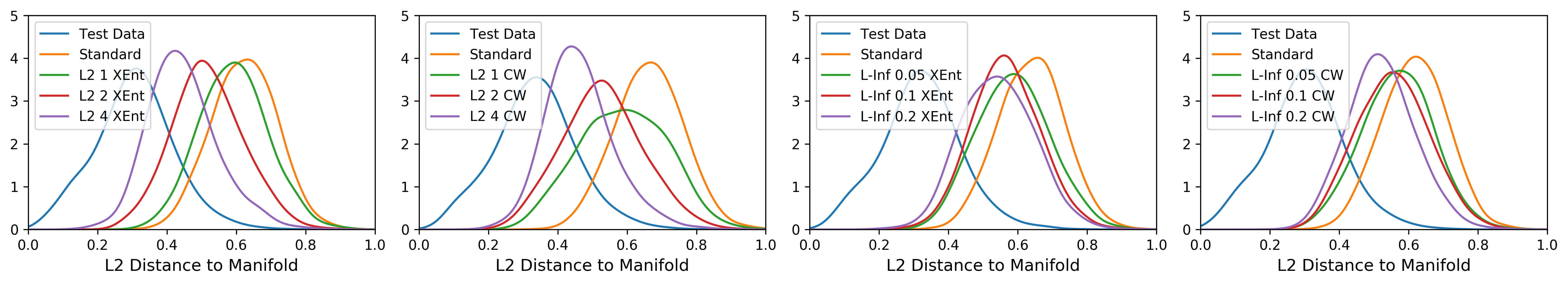}
	\caption{MNIST}
	\vspace{2mm}
	\end{subfigure}
	\begin{subfigure}{\linewidth}
	\includegraphics[width=\linewidth]{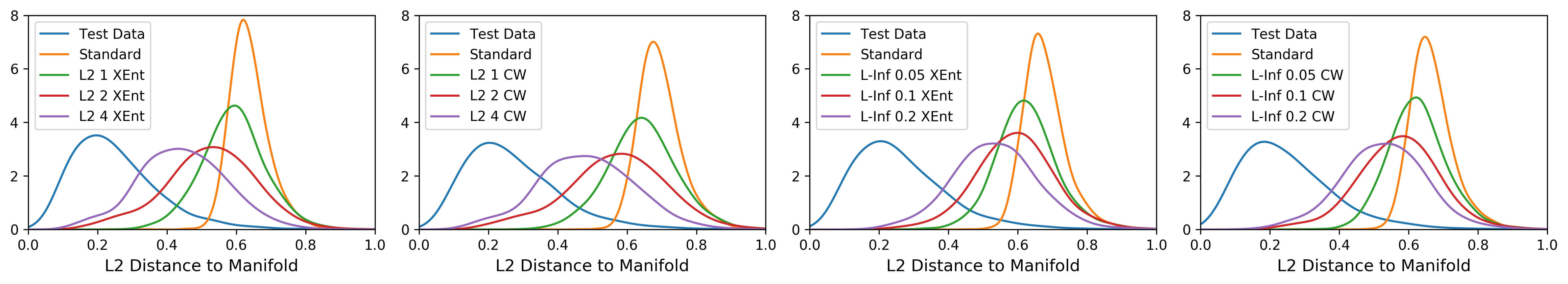}
	\caption{FMMNIST}
	\vspace{2mm}
	\end{subfigure}
	\begin{subfigure}{\linewidth}
	\includegraphics[width=\linewidth]{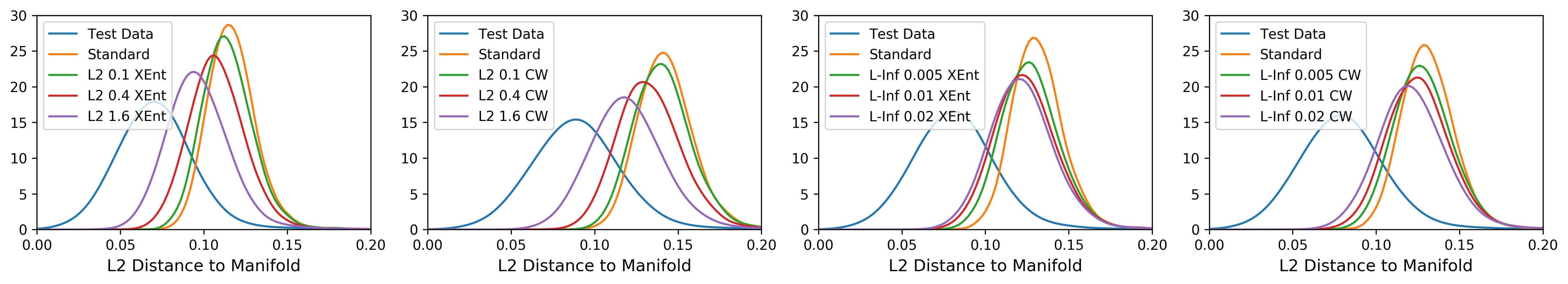}
	\caption{CIFAR-10}
	\end{subfigure}
	\caption{Distributions of $d_\pi$ of adversarial examples for standard and adversarially trained DNNs. We also show distributions of $d_\pi$ of test images for reference. The subcaptions indicate the dataset used. The adversaries that the DNNs trained against are denoted in legend by \textit{norm}, $\epsilon$ and \textit{objective}. Best viewed in electronic format (zoomed in).}
	\label{fig:gradient distance}
\end{figure}

Figure \ref{fig:gradient distance} compares distributions of $d_\pi$ for the test images and their adversarial examples from standard or adversarially trained DNNs. We only analyzed $\ell_2$-bounded attacks maximizing the XEnt loss since $\ell_2$-bounded attacks represent the original direction of the loss gradient while $\ell_\infty$-bounded attacks modify the gradient through clipping. Attacks which failed to change the prediction are removed since they are likely to be zero matrices due loss function saturation.

\begin{figure}[t]
	\centering
	\includegraphics[width=0.55\linewidth]{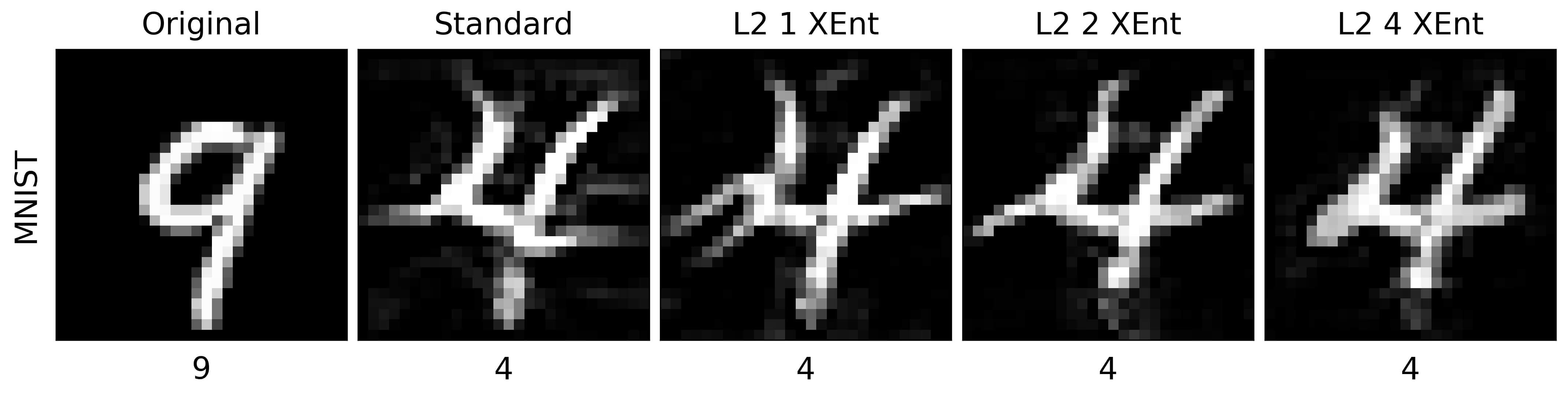}
	\includegraphics[width=0.55\linewidth]{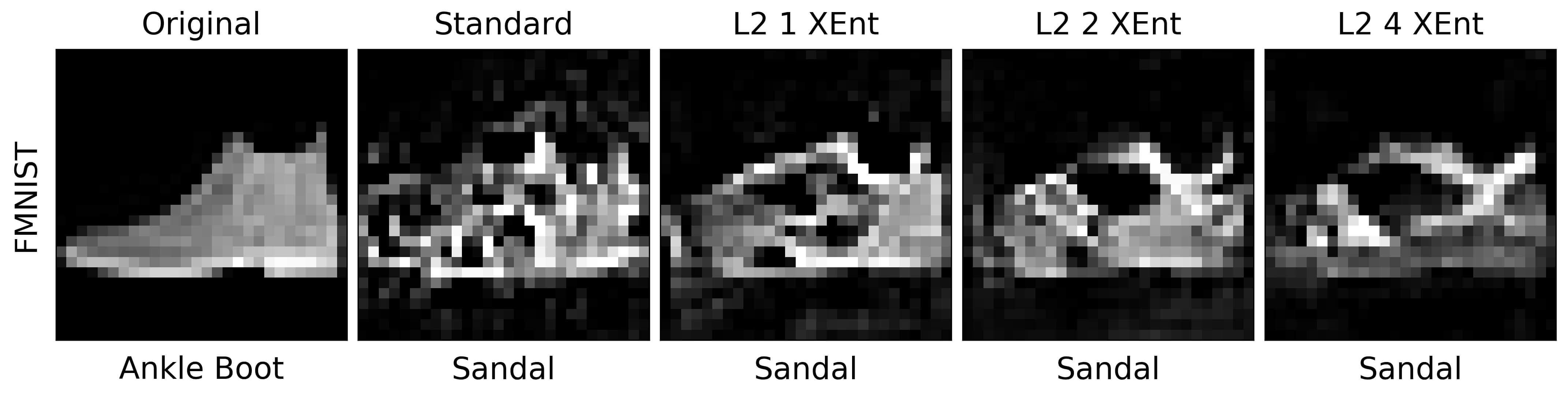}
	\includegraphics[width=0.55\linewidth]{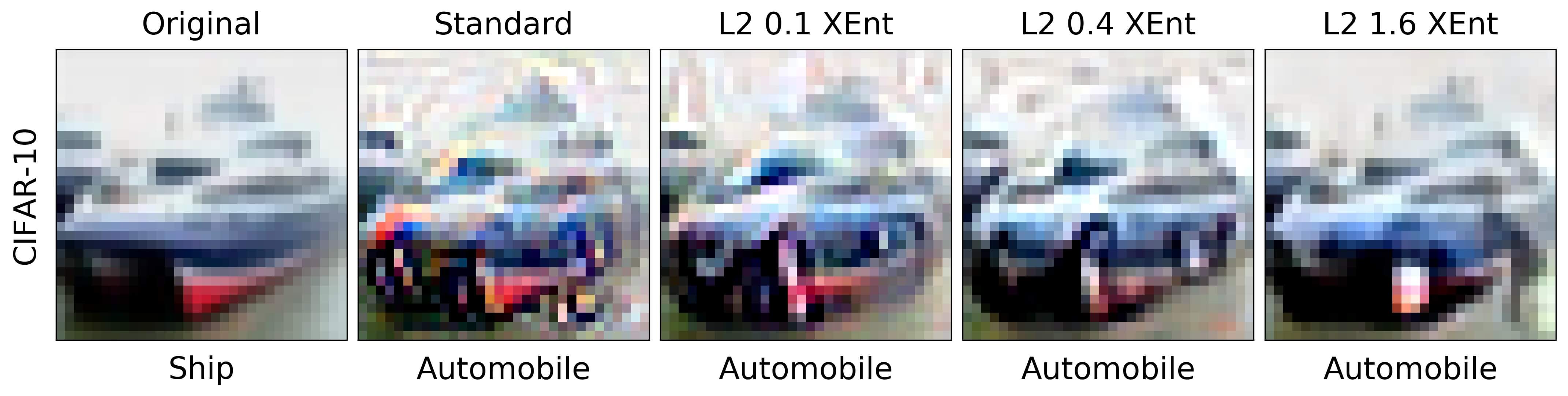}
	\caption{Visualization of adversarial examples for standard and adversarially trained DNNs. The adversaries that the DNNs trained against are denoted in the figure title by \textit{norm}, $\epsilon$ and \textit{objective}.}
	\label{fig:attacks}
\end{figure}

It can be observed from Figure \ref{fig:gradient distance} that, for all datasets, adversarial examples for adversarially trained DNNs lie closer to the image manifold than those for standard DNNs. This suggests  adversarial training restricts loss gradients to the image manifold. Hence gradients from adversarially trained DNNs are more visually interpretable. Interestingly, it can also be observed that using stronger attacks during training reduces the distance between adversarial examples and their projections even further. That is, the adversarial examples from more robust DNNs look more natural, as shown in Figure \ref{fig:attacks}. We now provide a conjecture for these phenomena.

\subsection{A Conjecture Under the Boundary Tilting Perspective} \label{section:visual sub2}

\begin{figure}[t]
	\centering
	\begin{subfigure}{0.41\linewidth}
	\includegraphics[width=\linewidth]{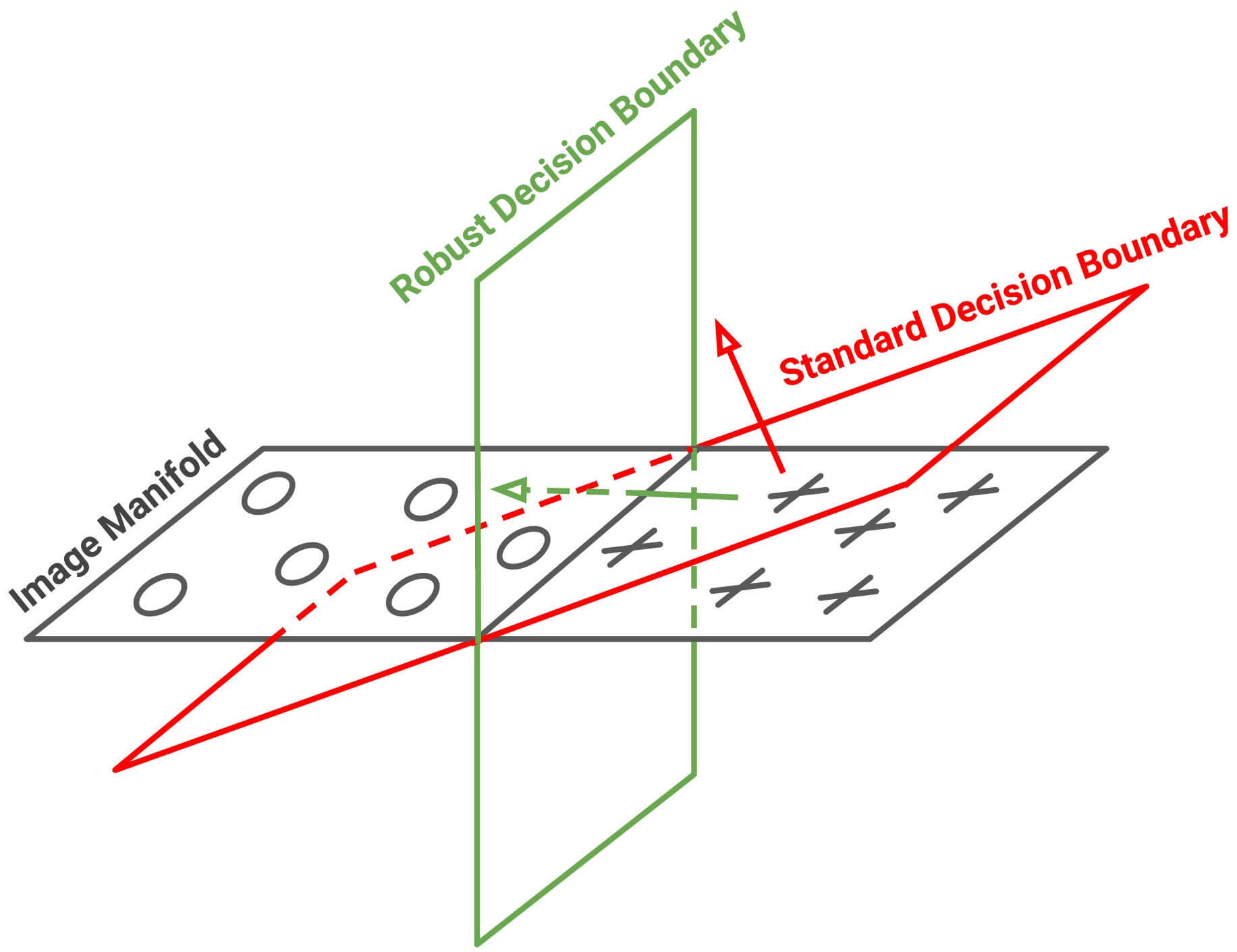}
	\caption{Illustration of the boundary tilting perspective.}
	\label{fig:boundary theory}
	\end{subfigure}
	\hspace{10mm}
	\begin{subfigure}{0.39\linewidth}
	\includegraphics[width=\linewidth]{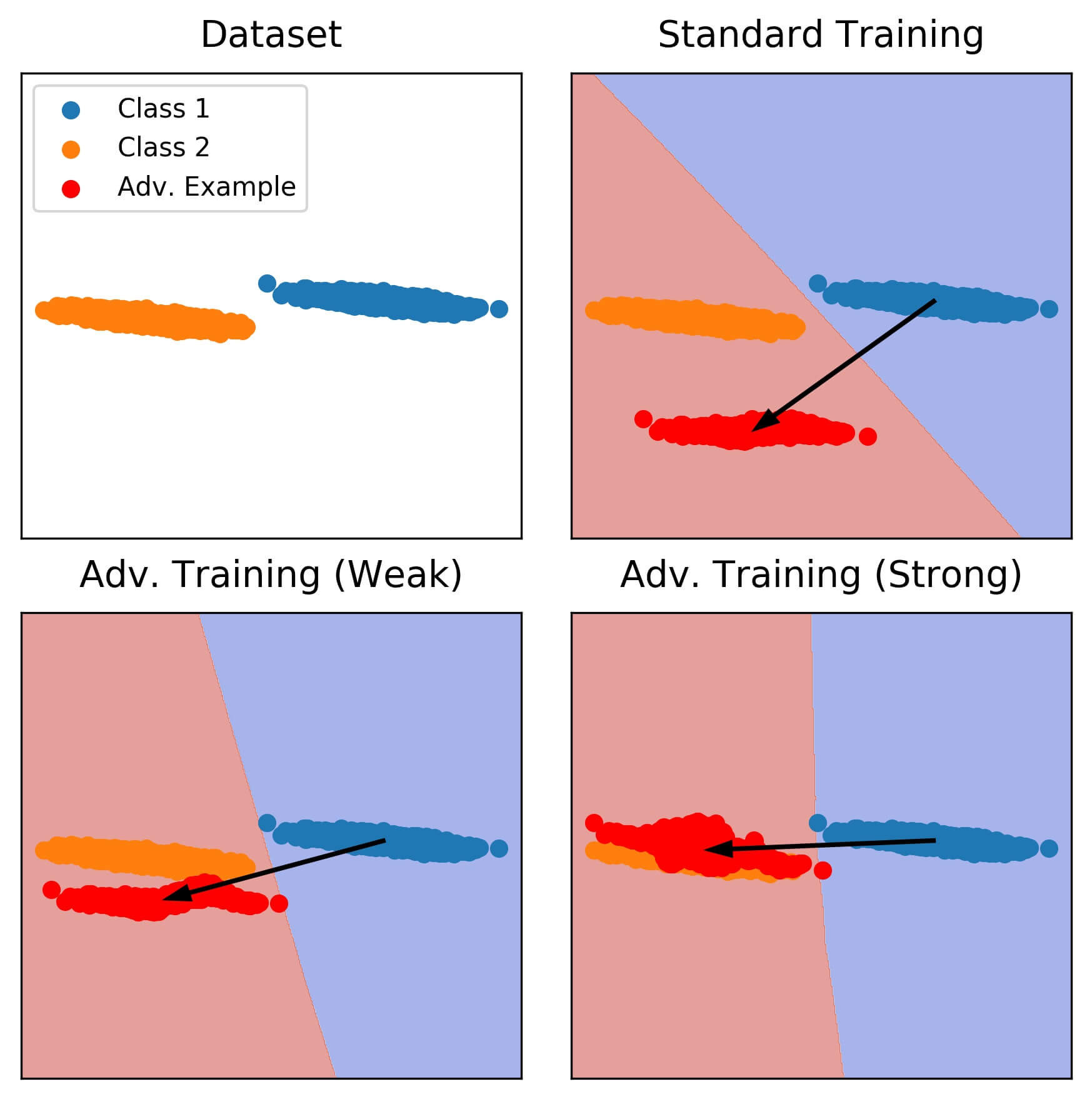}
	\caption{Results on a 2-dimensional toy dataset.}
	\label{fig:boundary exp}
	\end{subfigure}
	\caption{(a) An illustration of the boundary tilting perspective. Each arrow indicates adversarial perturbation against the decision boundary of corresponding color. (b) Experiment results on a 2-dimensional toy dataset. The blue and red regions indicate points where the network classified as class 1 and class 2. The adversarial attacks are taken against points in class 2. The bottom left and right figures show the instances where the network is trained against a weak and strong adversary.}
	\label{fig:conjecture}
\end{figure}

In this section, we propose a conjecture for why adversarial training confines the gradient to the manifold. Our conjecture is based on the boundary tilting perspective on the phenomenon of adversarial examples \citep{Tanay2016}. Figure \ref{fig:boundary theory} illustrates the boundary tilting perspective.
Adversarial examples exist because the decision boundary of a standard DNN is \enquote{tilted} along directions of low variance in the data (standard decision boundary). Under certain situations, the decision boundary will lie close to the data such that a small perturbation directed toward the boundary will cause the data to cross the boundary. Moreover, since the decision boundary is tilted, the perturbed data will leave the image manifold.

Under the boundary tilting perspective and observations of Section \ref{section:visual sub1}, we present a new conjecture on the relationship between adversarial training and visual interpretability of loss gradients: adversarial training removes tilting along directions of low variance in the data (robust decision boundary). Intuitively, this makes sense because a network is robust when only large-$\epsilon$ attacks are able to cause a nontrivial drop in accuracy, and this happens when data points of different classes are mirror images of each other with respect to the decision boundary. Since the loss gradient is generally perpendicular to the decision boundary, it will be confined to the image manifold and thus adversarial examples stay within the image manifold.

As a sanity check, we tested our conjecture with a 2-dimensional toy dataset. Specifically, we trained three two-layer ReLU network to classify points from two distinct bivariate Gaussian distributions. The first network is trained on original data and the latter two networks are trained against weak and strong adversaries. We then compared the resulting decision boundaries and the distribution of adversarial examples. Specific adversarial training procedures are described in Appendix \ref{section:2.2 settings}.

The training data and the results are shown in Figure \ref{fig:boundary exp}. The decision boundary of the standard network is tilted along directions of low variance in the data. Naturally, the adversarial examples leave the data manifold. On the other hand, adversarial training removes the tilt. Hence adversarial perturbations move data points to the manifold of the other class.
We also observed training against a stronger adversary removes the tilt to a larger degree. This causes adversarial examples to align better with the data manifold. Hence the decision boundary tilting perspective may also account for why adversarial training with stronger attack reduces the distance between an adversarial example and its projection even further (Figure \ref{fig:gradient distance}). We leave more theoretical justification and deeper experimental validation of our observations and hypothesis for future work.

If our conjecture is true, adversarial training prevents the decision boundary from tilting. Hence adversarial examples are restricted to the image manifold and thus loss gradients align better with human perception. However, as we have discussed in Section \ref{section:introduction}, visual sharpness of gradients do not imply that they accurately represent the features learned by the DNN. We address this issue in the next section by quantitatively evaluating the interpretability of loss gradients.

\section{Interpretability of Adversarially Trained Networks} \label{section:quantitative}

To the best of our knowledge, there are no works on quantifying the interpretability of loss gradients. However, quantitative interpretability of logit gradients and its variants have been thoroughly studied in the context of attribution methods \citep{Bach2015,Samek2017,Adebayo2018,Hooker2018}.\footnote{Attribution methods are DNN decision interpretation methods which assign a signed \textit{attribution} value to each input feature (pixel). An attribution value quantifies the amount of influence of the corresponding feature on the final decision. Since each pixel is assigned an attribution value, we can visualize the attributions by arranging them to have the same shape as the input.} Since the loss gradient highlights the input features which affect the loss most strongly, and thus are important for the DNN's prediction, we may also treat it as an attribution method. This allows us to extend the reasoning and techniques in works on attribution method evaluation to the loss gradient.\footnote{Since the loss gradient is a linear combination of logit gradients, we can reasonably expect most observations for loss gradients to hold for logit gradients as well (and vice versa). However, loss gradients have the nice property of being generally perpendicular to the decision boundary, and this gives us more insights such as Section \ref{section:visual sub2}. Hence we have chosen to examine loss gradients, not logit gradients.}


\subsection{Formal Description of the Quantitative Evaluation Framework} \label{section:quantitative sub1}

We denote vectors or vector-valued functions by boldface letters. Let $\mathcal{F}, \mathcal{G}$ and $ \mathcal{M}$ denote families of DNNs, attribution methods and attribution method evaluation metrics. A DNN $\mathbf{f} \in \mathcal{F}$ maps an image $\mathbf{x} \in \mathbb{R}^d$ to a vector of class logits $\mathbf{y} \in \mathbb{R}^C$, where $d$ is the input dimension and $C$ is the number of classes. Then an attribution method $\mathbf{g} \in \mathcal{G}$ maps the tuple $(\mathbf{f}, \mathbf{x})$ to a vector of attribution scores $\mathbf{a} \in \mathbb{R}^d$. Finally, an attribution method evaluation metric $\mu \in \mathcal{M}$ assigns to each tuple $(\mathbf{f}, \mathbf{g})$ a scalar indicating how accurately $\mathbf{g}$ reflects the internal representation of $\mathbf{f}$. A higher value of $\mu$ indicates that the attribution method better describes the internal representation of the DNN.

Here, note that the value of $\mu$ depends on both $\mathbf{f}$ and $\mathbf{g}$. Previous works have focused on improving the value of $\mu$ for fixed $\mathbf{f}$ by developing better and more complex $\mathbf{g}$ \citep{Bach2015,Sundararajan2017,Smilkov2017,Shrikumar2017}. However, we can also improve the value of $\mu$ for fixed $\mathbf{g}$ by varying $\mathbf{f}$ through changing the network topology, training scheme, etc. It is only recently that the latter approach have started to receive attention. In the next section, we investigate this approach in the context of adversarial training.

\subsection{Effect of Adversarial Training on Loss Gradient Interpretability} \label{section:quantitative sub2}

Here we experimentally evaluate whether loss gradients from adversarially trained DNNs are truly more interpretable than those from standard DNNs. Using the definitions from the previous section, we can rephrase this goal as follows: Let $\mathcal{F}_{Std}$ be the family of standard DNNs and let $\mathcal{F}_{Adv}$ be the family of adversarially trained DNNs, all of the same architecture. Given an attribution method $\mathbf{g} \in \mathcal{G}$, we want to verify whether $\min_{\mathbf{f} \in \mathcal{F}_{Adv}} \mu(\mathbf{f}, \mathbf{g}) > \max_{\mathbf{f} \in \mathcal{F}_{Std}} \mu(\mathbf{f}, \mathbf{g})$. In particular, we are interested in the case when $\mathbf{g}(\mathbf{f}, \mathbf{x}) = \nabla_{\mathbf{x}} J$ where $J$ is the XEnt loss. We denote this attribution method by $\mathbf{g}_{\,G}$. We also evaluate a variant Gradient $*$ Input \citep{Shrikumar2017}: $\mathbf{g}(\mathbf{f}, \mathbf{x}) = \mathbf{x} \odot \nabla_{\mathbf{x}} J$. We denote Gradient $*$ Input using the XEnt loss by $\mathbf{g}_{\,GX}$.

\begin{figure}[t]
	\centering
	\includegraphics[width=0.48\linewidth]{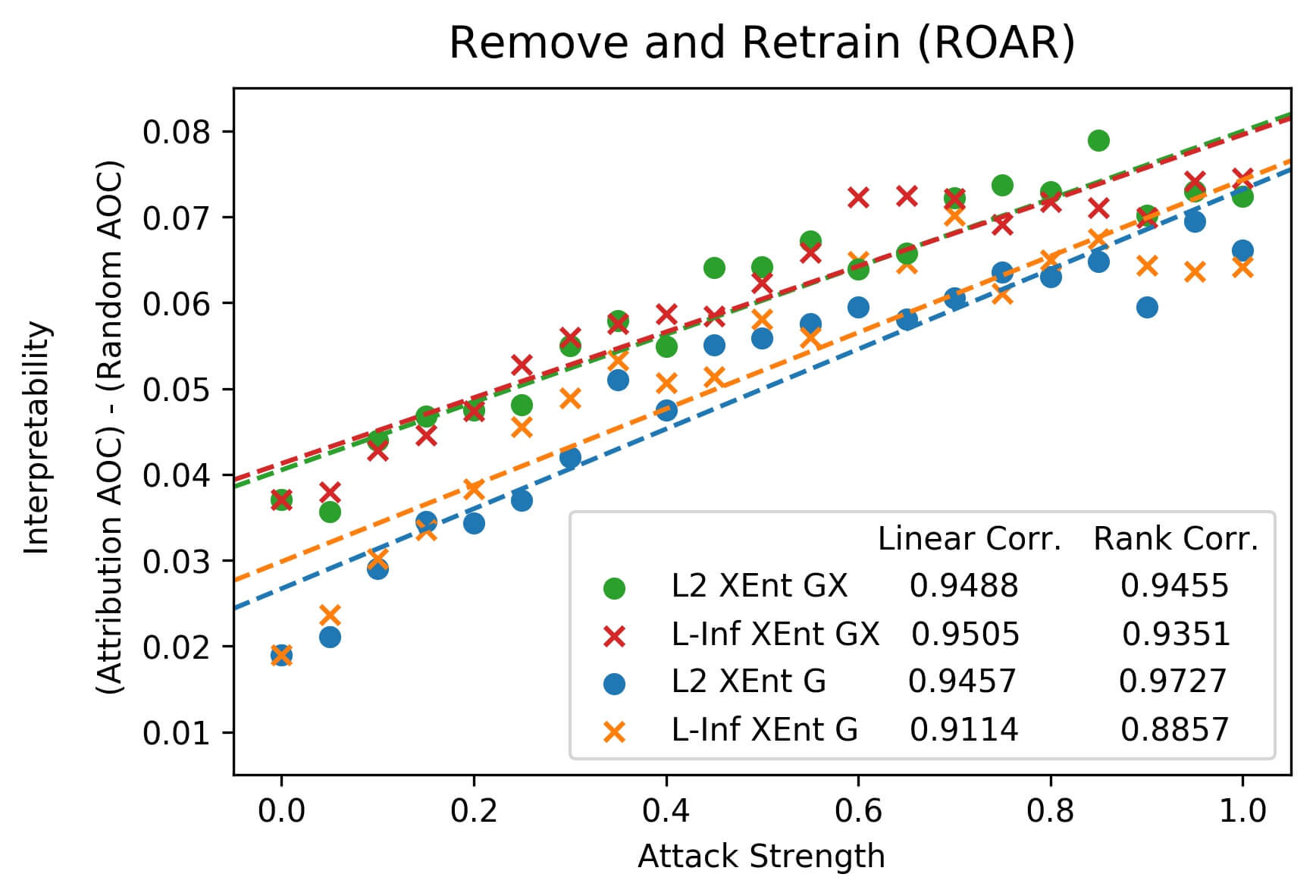}
	\includegraphics[width=0.485\linewidth]{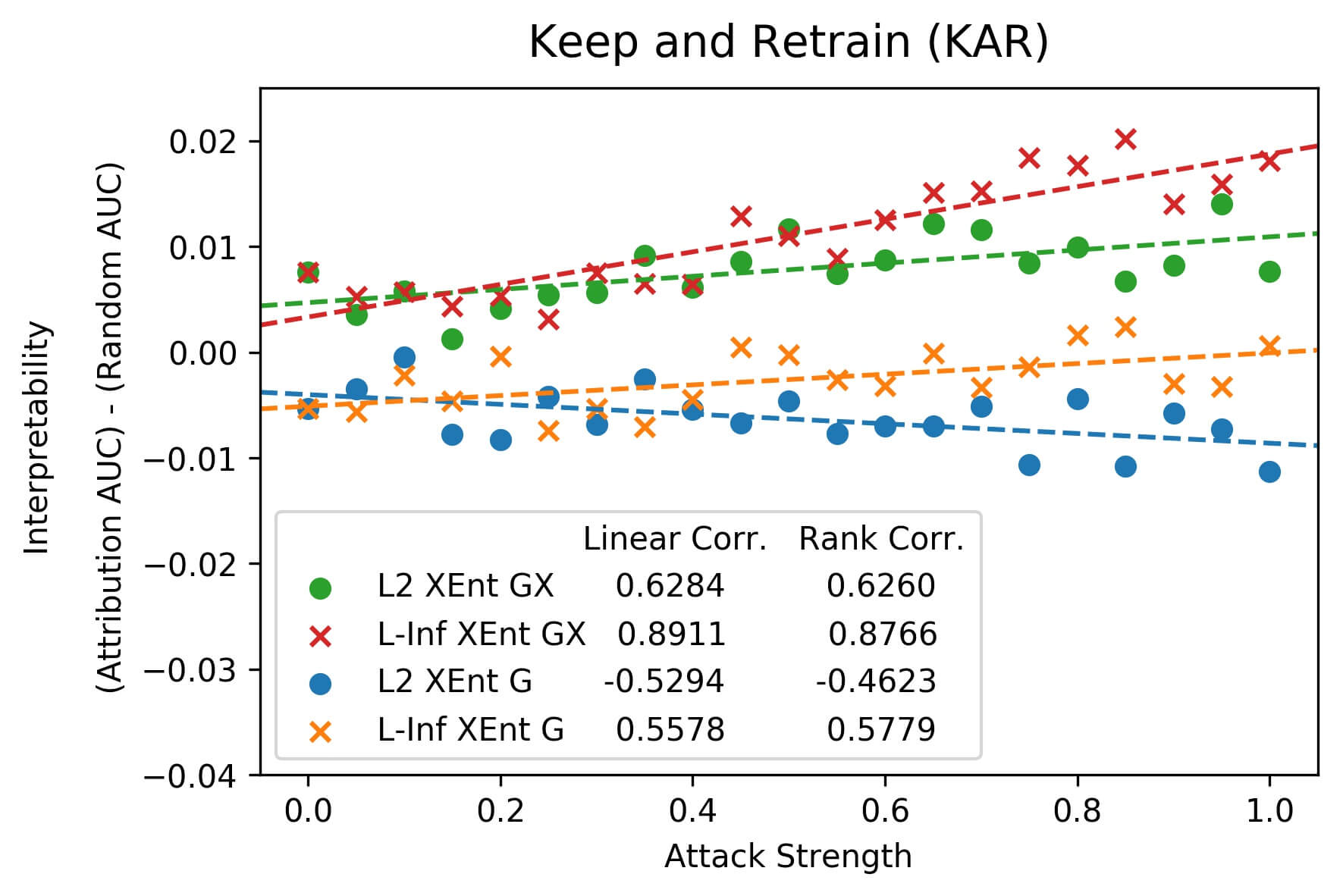}
	\includegraphics[width=0.48\linewidth]{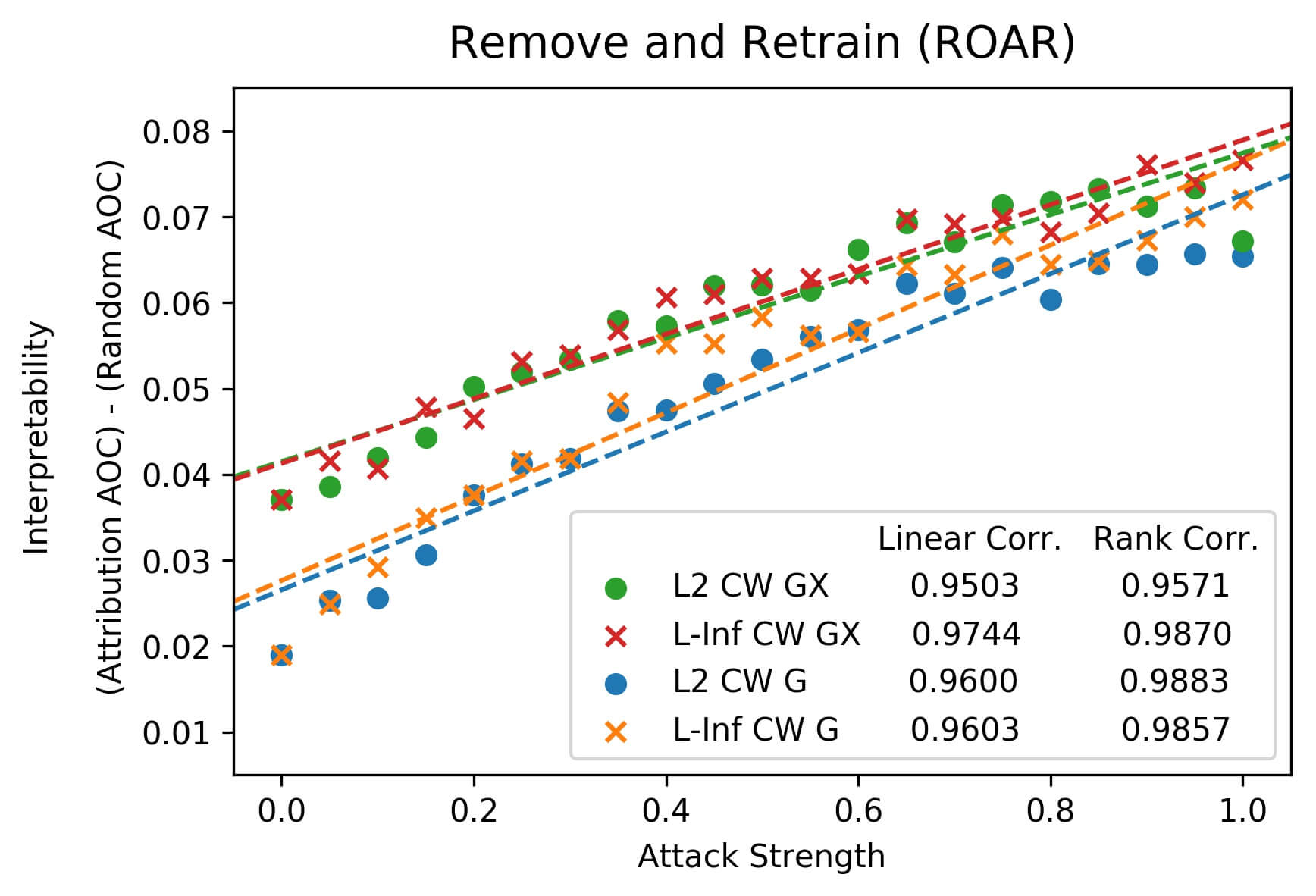}
	\includegraphics[width=0.485\linewidth]{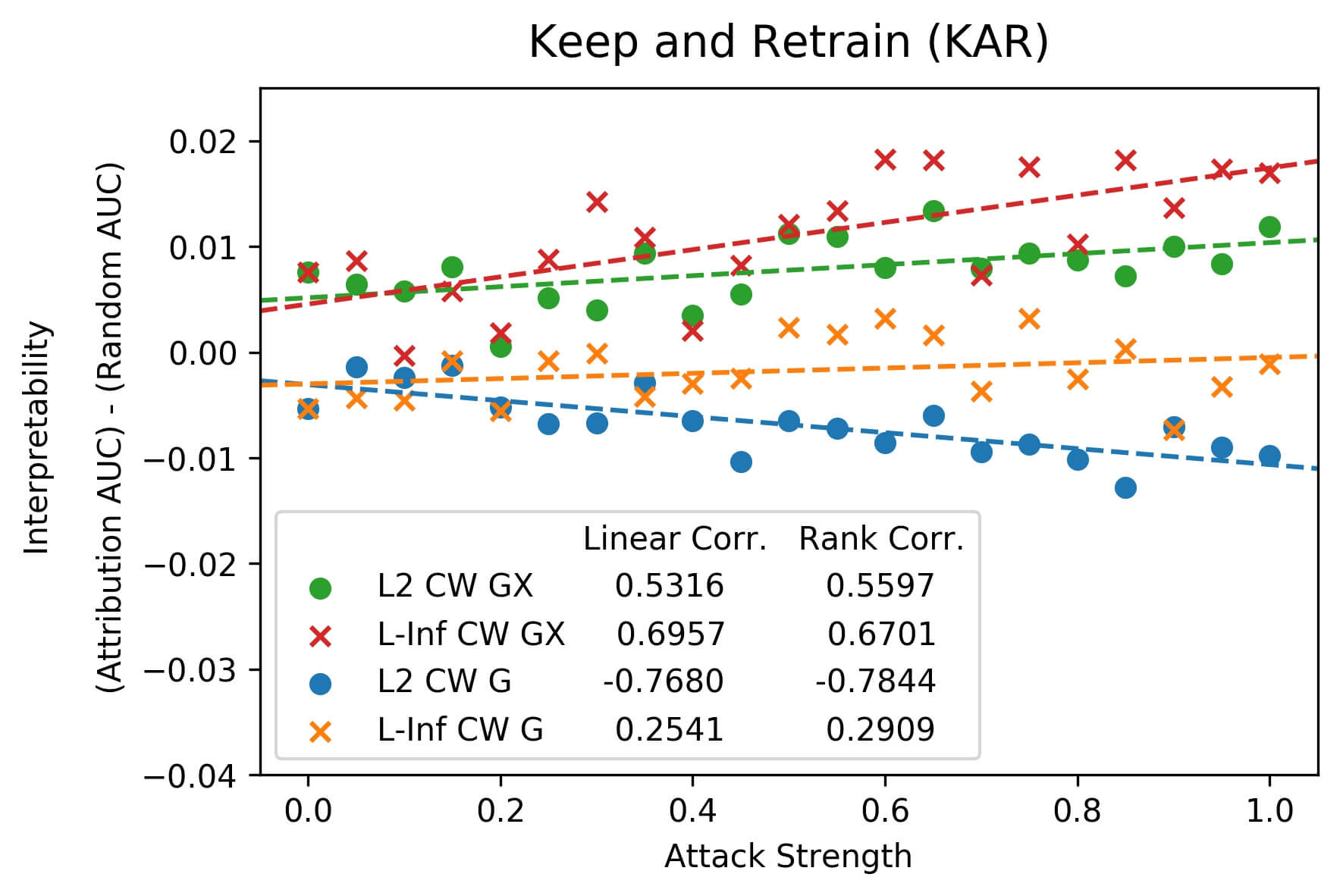}
	\caption{Effect of adversarial training on interpretability of $\mathbf{g}_{\,G}$ and $\mathbf{g}_{\,GX}$. The x-axis indicates $\epsilon$ (attack strength) used during training. The values of $\epsilon$ are scaled into $[0,1]$ such that $\ell_2$-bounded attacks and $\ell_\infty$-bounded attacks are comparable. Note that $\epsilon = 0$ is standard training. The y-axis indicates quantitative interpretability, as explained in the text. We also show the linear correlation coefficient and Spearman's rank correlation coefficient for each combination of adversarial training setting (\textit{norm} and \textit{objective}) and attribution method (G for $\mathbf{g}_{\,G}$ and GX for $\mathbf{g}_{\,GX}$).}
	\label{fig:roar kar}
\end{figure}

We quantified the interpretability of each methods via two attribution method evaluation metrics Remove and Retrain (ROAR) and Keep and Retrain (KAR) \citep{Hooker2018}. Specifically, we measured how the performance of the classifier changed as features were occluded based on the ordering assigned by the attribution method. For ROAR, we replaced a fraction of all CIFAR-10 pixels estimated to be \textit{most} important with a constant value. For KAR, we replaced the pixels estimated to be \textit{least} important. We then retrained a CNN on the modified dataset and measured the change in test accuracy. We trained three CNNs per attribution method for each fraction $\{0.1, 0.3, 0.5, 0.7, 0.9\}$ and measured test accuracy as the average of these three CNNs.

Since ROAR removes most important input features, a better attribution method should cause more accuracy degradation. Conversely, since KAR removes least important features, a better attribution method should cause less accuracy degradation. For reference, we also evaluated the random baseline $\mathbf{g}_{\,Rand}$ which assigns random attribution values to input features. We then defined the interpretability scores under ROAR and KAR by
\begin{align*}
\mu_{ROAR}(\mathbf{f}, \mathbf{g}) &= (\text{Area over curve (AOC) of the ROAR curve for } \mathbf{g}) - (\text{AOC for } \mathbf{g}_{\,Rand}), \\
\mu_{KAR}(\mathbf{f}, \mathbf{g}) &= (\text{Area under curve (AUC) of the KAR curve for } \mathbf{g}) - (\text{AUC for } \mathbf{g}_{\,Rand}).
\end{align*}

Figure \ref{fig:roar kar} shows the values of $\mu_{\,ROAR}$ and $\mu_{\,KAR}$ for each adversarial training setting. Specific adversarial training procedures are described in Appendix \ref{section:3 settings}. First, we observed that there is a strong positive correlation between the strength of attack used during adversarial training and interpretability with the exception of $\mathbf{g}_{\,G}$ from $\ell_2$-trained DNNs in KAR. This result is significant because it shows adversarial training indeed causes the gradient to better reflect the internal representation of the DNN. It implies that training with an appropriate \enquote{interpretability regularizer} may be enough to produce DNNs that can be interpreted with simple attribution methods such as gradient or Gradient $*$ Input.
However, it does \emph{not} imply we no longer need to develop complex attribution methods to interpret DNNs. This topic will be dealt with further in Section \ref{section:quantitative sub3}, in the context of trade-off between accuracy and loss gradient interpretability.

We also observed Gradient $*$ Input performs better than the loss gradient. We believe this is because the former method is a global attribution method while the latter is a local attribution method.\footnote{Local attribution methods return vectors that maximize the value which the attribution is taken with respect to. On the other hand, global attribution methods visualize the marginal effect of each feature on that value. For further details on the difference between local and global attribution methods, we refer the readers to Section 3.2 of \citet{Ancona2017}.} Since both ROAR and KAR evaluate attribution methods based on feature occlusion, Gradient $*$ Input should theoretically show better performance.

Finally, we remark that if our conjecture in Section \ref{section:visual sub2} is true, there may be a close connection between gradient interpretability and the degree to which gradient is confined to the data manifold. In other words, DNNs with less tilted decision boundaries may yield loss gradients that are more visually and quantitatively meaningful.

\subsection{Accuracy and Loss Gradient Interpretability Trade-off} \label{section:quantitative sub3}

Previous works have observed that there may be a trade-off between accuracy and adversarial robustness \citep{Tsipras2018,Su2018}. As we have shown in the previous section, there exists a positive correlation between the strength of adversarial attack used in the training process and gradient interpretability. Hence it is highly likely that \textit{there exists a negative correlation between network accuracy and gradient interpretability}. To verify this, we trained CNNs on CIFAR-10 under various adversarial attack settings and evaluated their gradient interpretability. More detailed experiments setting used in this subsection can be found in Appendix \ref{section:3 settings}.

\begin{figure}[t]
	\centering
	\includegraphics[width=0.48\linewidth]{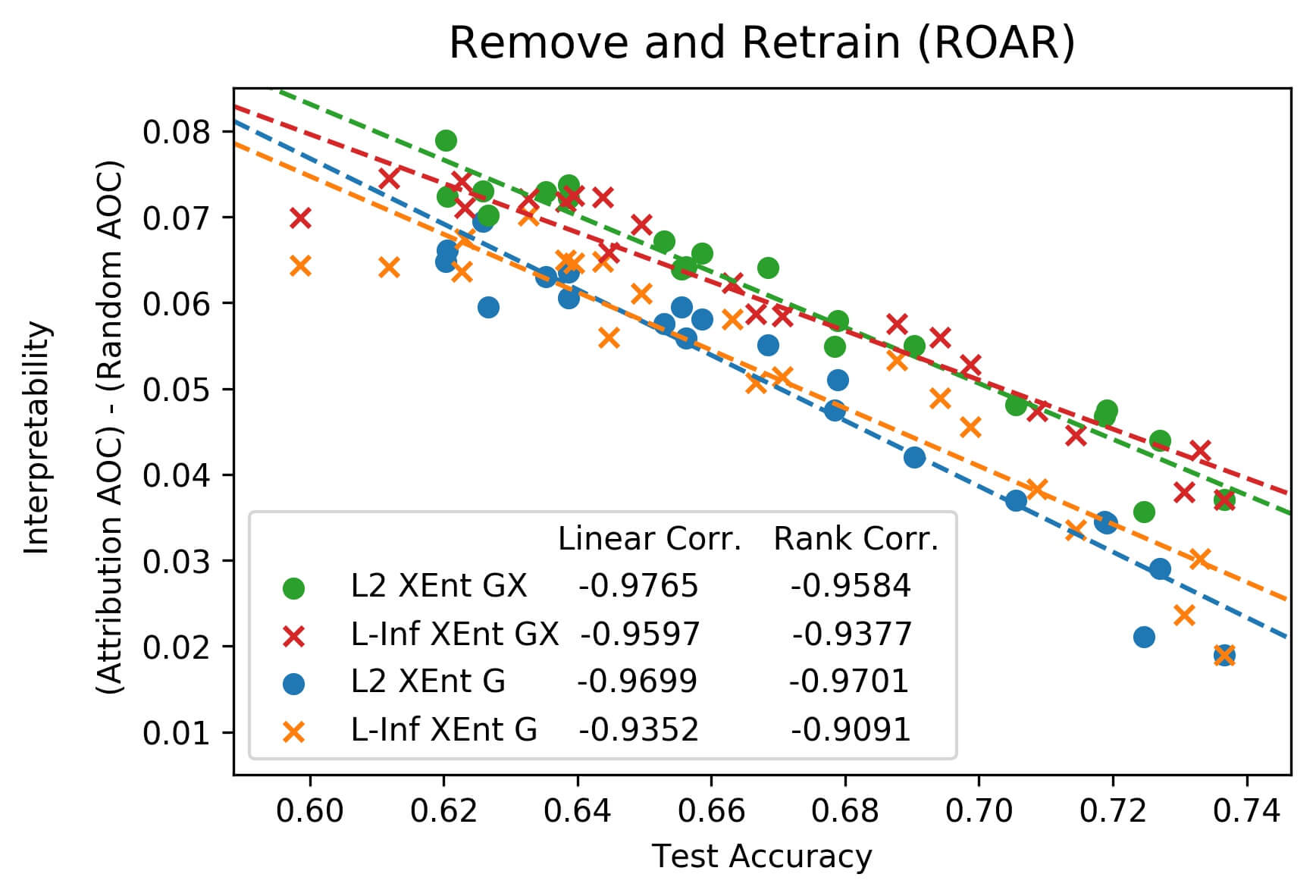}
	\includegraphics[width=0.485\linewidth]{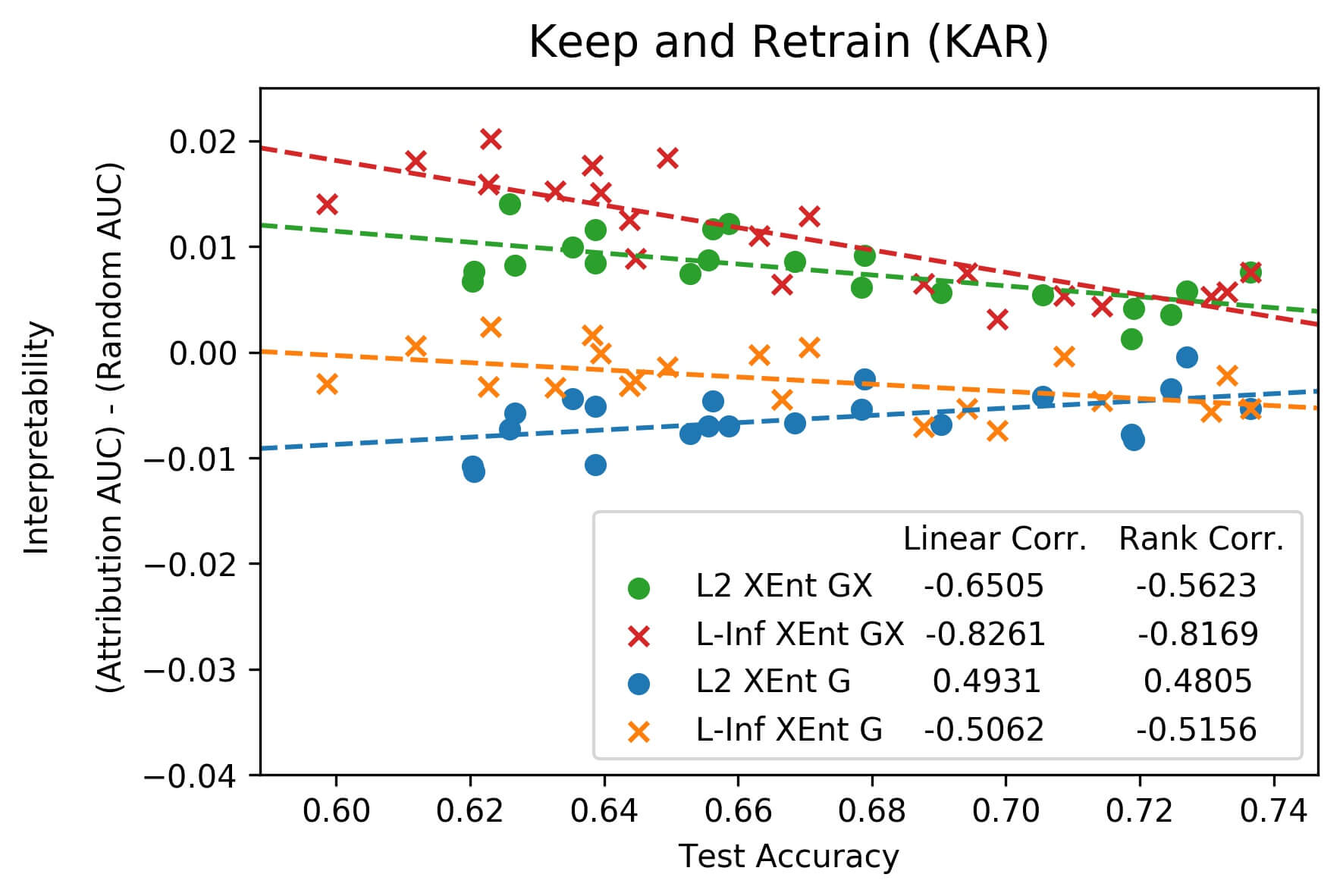}
	\includegraphics[width=0.48\linewidth]{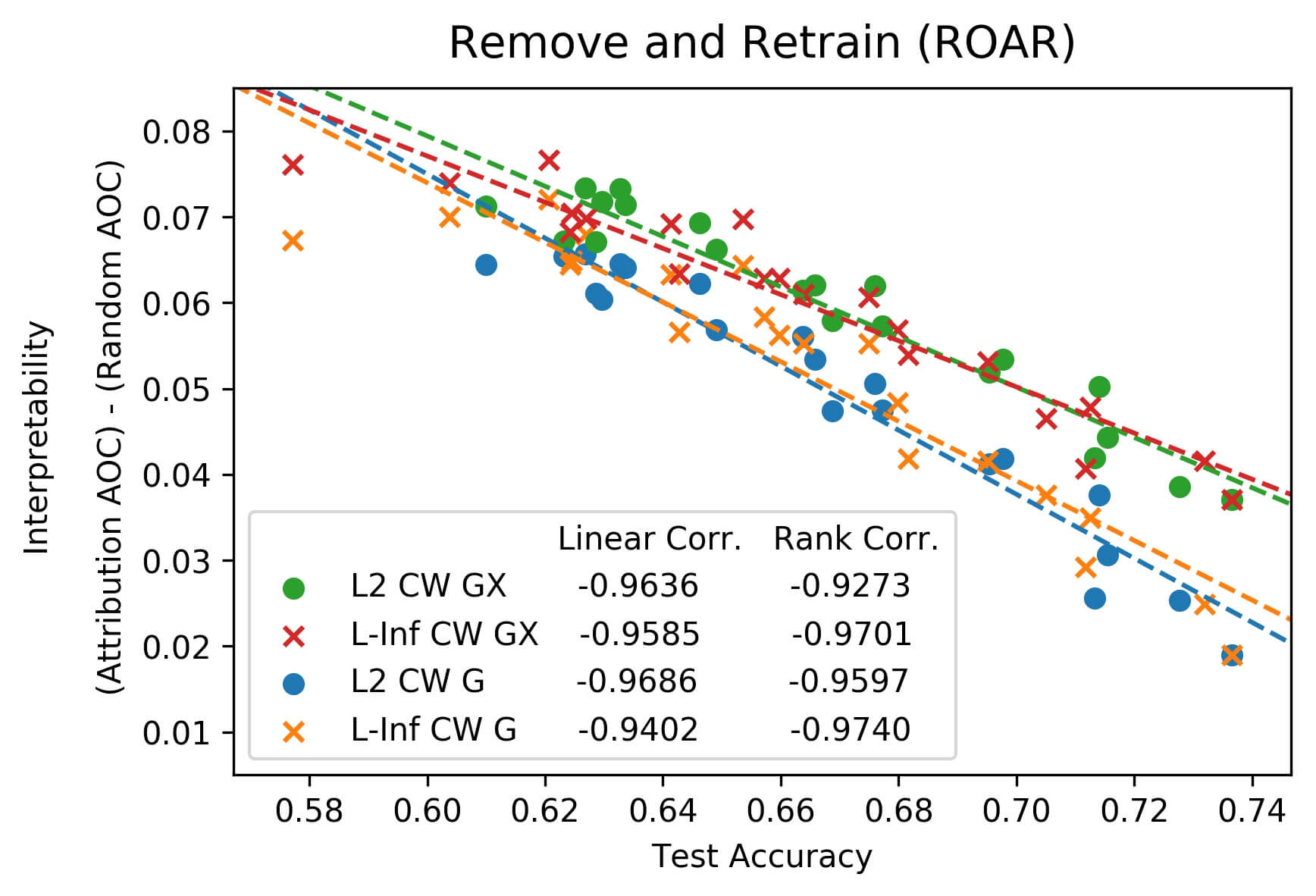}
	\includegraphics[width=0.485\linewidth]{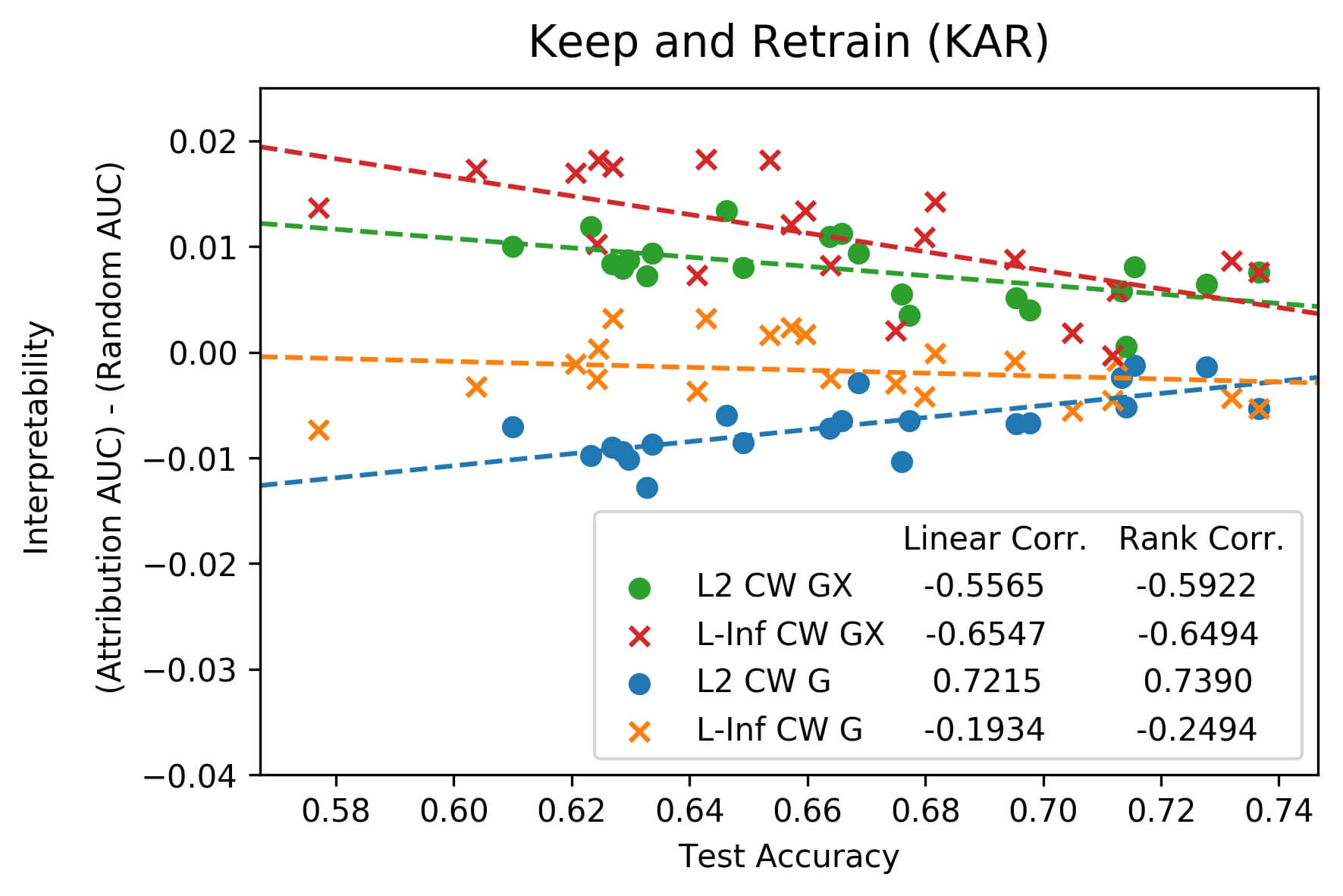}
	\caption{Relation between test accuracy and interpretability of $\mathbf{g}_{\,G}$ and $\mathbf{g}_{\,GX}$ under the adversarial training framework. The x-axis indicates test accuracy on natural images. The y-axis indicates quantitative interpretability, as explained in the text. We also show the linear correlation coefficient and Spearman's rank correlation coefficient for each combination of adversarial training setting (\textit{norm} and \textit{objective}) and attribution method (G for $\mathbf{g}_{\,G}$ and GX for $\mathbf{g}_{\,GX}$).}
	\label{fig:roar kar tradeoff}
\end{figure}

\begin{figure}[t]
	\centering
	\begin{subfigure}{0.29\linewidth}
	\includegraphics[width=\linewidth]{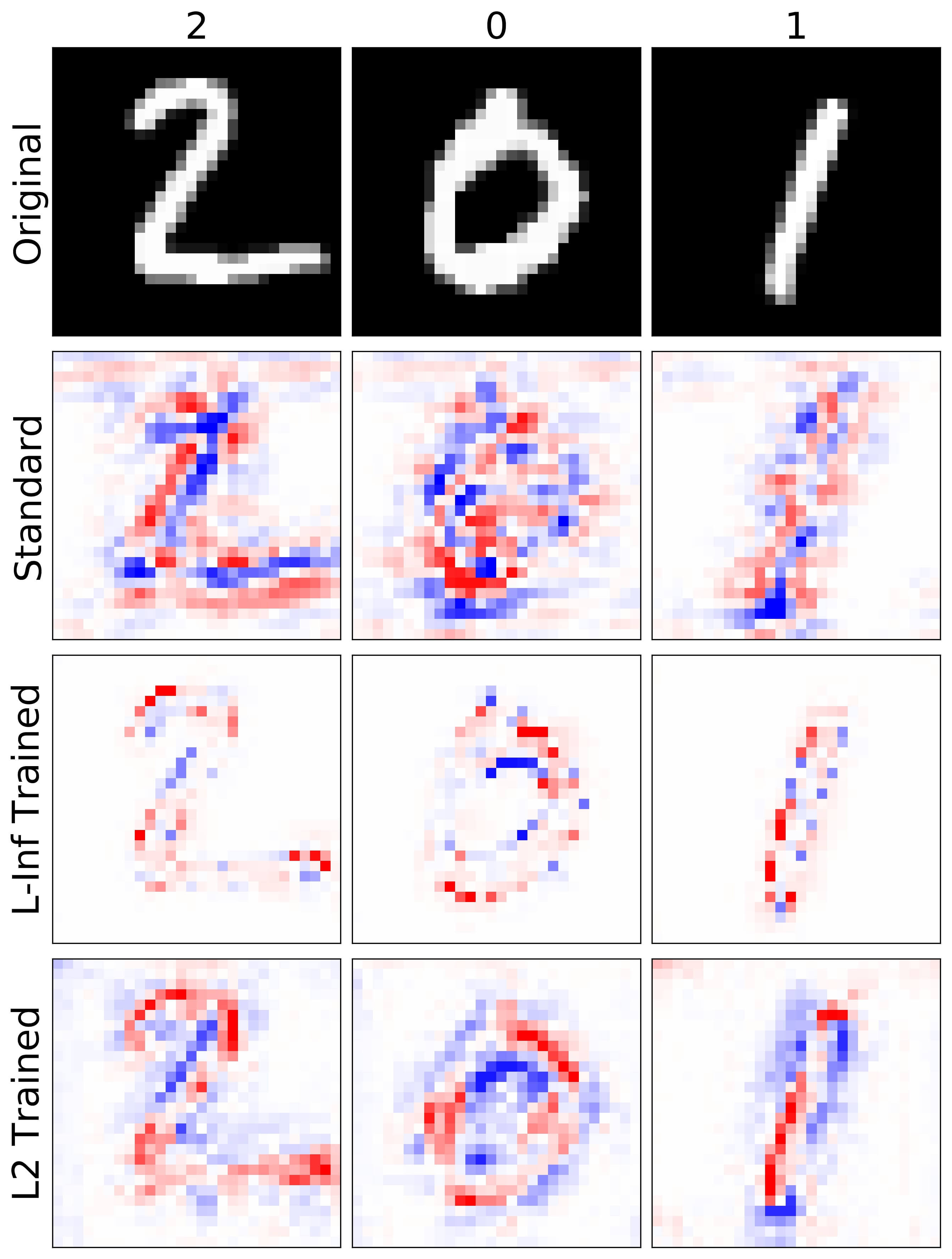}
	\caption{MNIST}
	\end{subfigure}
	\hfill
	\begin{subfigure}{0.29\linewidth}
	\includegraphics[width=\linewidth]{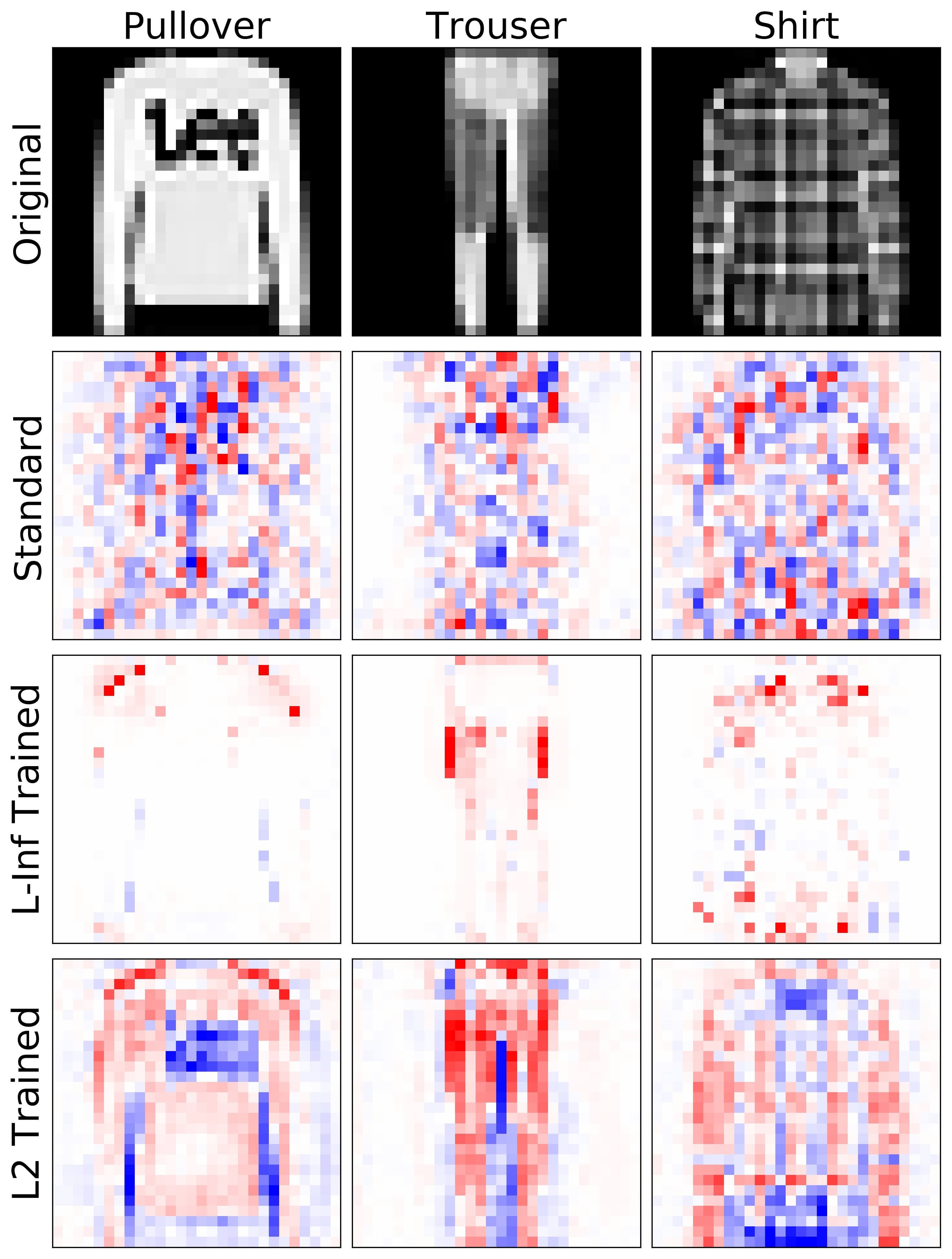}
	\caption{FMNIST}
	\end{subfigure}
	\hfill
	\begin{subfigure}{0.29\linewidth}
	\includegraphics[width=\linewidth]{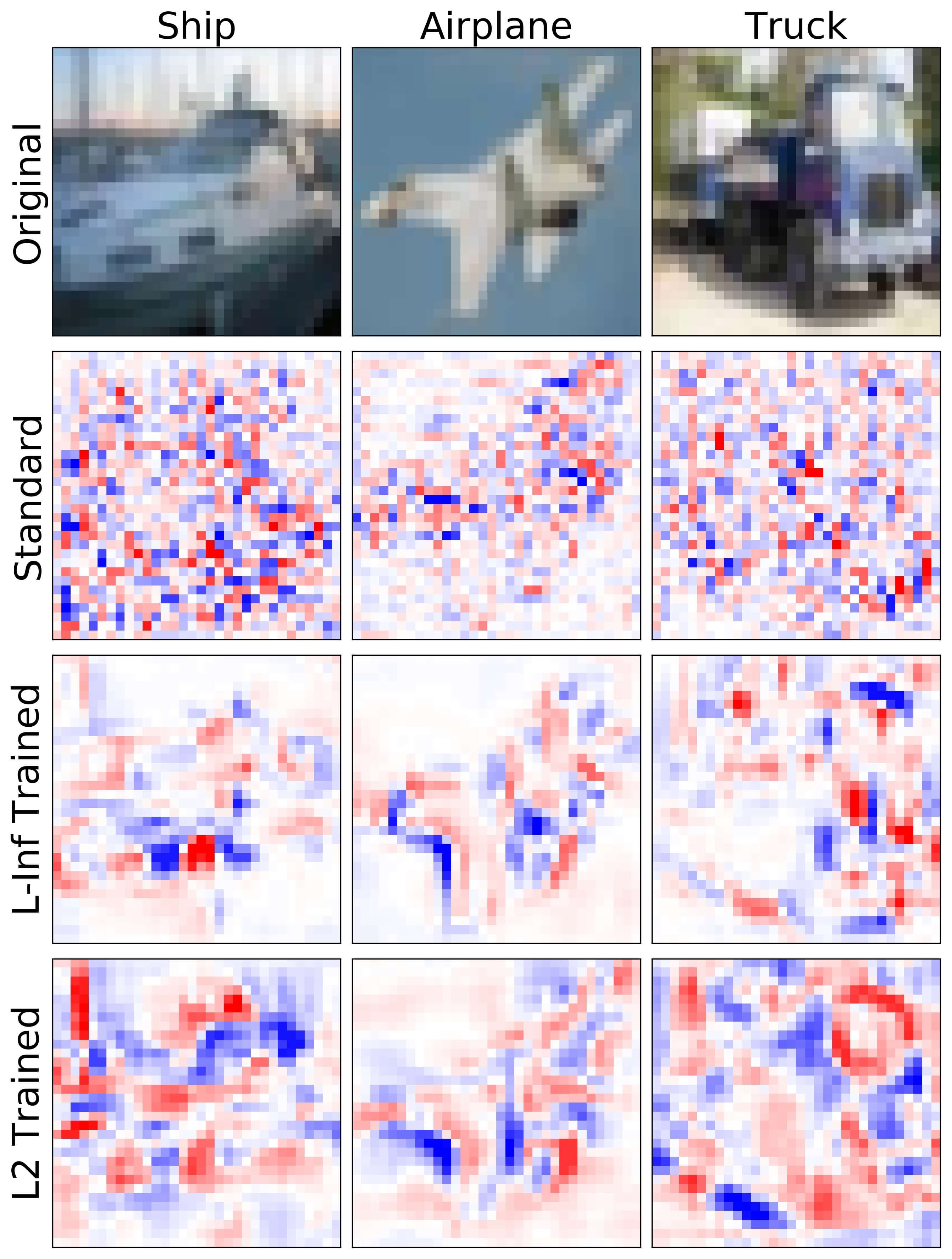}
	\caption{CIFAR-10}
	\end{subfigure}
	\caption{Visualization of the loss gradient for standard and adversarially trained DNNs. To display the gradients, we summed up the gradients along the color channel and then capped low outlying values to 0.5\textsuperscript{th} percentile and high outlying values to 99.5\textsuperscript{th} percentile.}
	\label{fig:loss gradient}
\end{figure}

Figure \ref{fig:roar kar tradeoff} shows the relation between test accuracy and loss gradient interpretability. Indeed, there is a near-monotonic decreasing relation between interpretability and accuracy under both ROAR and KAR. We note that the only exception of this trend is $\mathbf{g}_{\,G}$ from $\ell_2$-trained DNNs in KAR.
These results imply that adversarial training itself is \emph{not} a perfect method for attaining gradient interpretability without sacrificing test accuracy.

We also observed that attributions from $\ell_2$-trained networks are more resistant to this trade-off in ROAR. On the other hand, attributions from $\ell_\infty$-trained networks were more resistant in KAR. This implies attributions from $\ell_2$-trained DNNs are more effective at emphasizing important features while attributions from $\ell_\infty$-trained DNNs are better at identifying less important features. This is somewhat consistent with the visual characteristics of loss gradients: as shown in Figure \ref{fig:loss gradient}, gradients from $\ell_\infty$-trained networks are very sparse but discontinuous while gradients from $\ell_2$-trained networks are smooth but less sparse. Analyzing the effect of norm used to constrain the adversary on the neural network's decision boundary and the gradient may also be an interesting line of research.

From the results, we see two potential approaches to resolving this trade-off. First, since the global attribution method $\mathbf{g}_{\,GX}$ performs better than the local attribution method $\mathbf{g}_{\,G}$, we can explore combinations of adversarial training with other global attribution methods such as Layer-wise Relevance Propagation \citep{Bach2015},  DeepLIFT \citep{Shrikumar2017} or Integrated Gradient \citep{Sundararajan2017}. Second, since there is large performance gain in using $\ell_\infty$-training over $\ell_2$-training in KAR while there is only slight gain in using $\ell_2$-training over $\ell_\infty$-training in ROAR, we can seek better ways of applying $\ell_\infty$-training.

\section{Conclusion} \label{section:conclusion}

Adversarial training is a training scheme designed to counter adversarial attacks by augmenting the training dataset with adversarial examples. Surprisingly, several studies have observed that loss gradients from adversarially trained DNNs are visually more interpretable than those from DNNs trained on natural images. Although this phenomenon is interesting, there are only few works that have offered an explanation. In this paper, we attempted to bridge this gap between adversarial robustness and gradient interpretability. To this end, we identified that loss gradients from adversarially trained DNNs align better with human perception because adversarial training restricts loss gradients closer to the image manifold. We also provided a conjecture for this phenomenon and verified its plausibility with a toy dataset. We then demonstrated that adversarial training indeed causes gradients to be quantitatively meaningful with two attribution method evaluation metrics. Finally, we showed with CNNs trained on CIFAR-10 that under the adversarial training framework, there exists an empirical trade-off between test accuracy and gradient interpretability. Then based on the experiment results, we proposed two potential approaches to resolving this trade-off.

\newpage

\bibliography{iclr2019_conference}
\bibliographystyle{iclr2019_conference}

\appendix

\newpage

\section{Experiment Settings} \label{section:settings}

\subsection{Datasets} \label{section:settings dataset}

The toy dataset is comprised two classes. Each class consists of 3000 points sampled from a bivariate Gaussian distribution. The first distribution has mean and covariance matrix
\begin{align*}
\mu =
\begin{bmatrix}
1.2 \\
0.1
\end{bmatrix}, \quad
\Sigma =
\begin{bmatrix}
\phantom{-} 0.1 & -0.01 \\
-0.01 & \phantom{-} 0.0002
\end{bmatrix}.
\end{align*}
The second distribution has $\mu = [1.2, 0.1]^T$ and the same covariance matrix. All four datasets toy dataset, MNIST, FMNIST and CIFAR-10 were normalized into range $[-1,1]$.

\subsection{Classification Networks} \label{section:settings classifiers}

We trained all the models with Adam with default settings $\alpha = 0.001$, $\beta_1 = 0.9$ and $\beta_2 = 0.999$. For the toy dataset, we trained a two-layer ReLU DNN for 10 epochs to achieve $100\%$ test accuracy. For MNIST and FMNIST, we trained a ReLU CNN for 5 epochs to achieve $98.6\%$ and $90.6\%$ test accuracy, respectively. For CIFAR-10, we trained a ReLU CNN for 20 epochs to achieve $73.6\%$ test accuracy. The architectures for the classification models are given in the tables below. For dense layers, we write \enquote{Dense (\textit{number of units})}. For convolution layers, we write \enquote{Conv 2D (\textit{filter size}, \textit{stride}, \textit{number of filters})}. For max-pooling layers, we write \enquote{Max-pooling (\textit{window size}, \textit{stride})}.

\begin{center}
\begin{tabular}{|c|}
\hline
\textbf{Toy Dataset DNN} \\
\hline
Dense (128) ReLU \\
\hline
Dense (2) \\
\hline
\end{tabular}
\quad
\begin{tabular}{|c|}
\hline
\textbf{MNIST / FMNIST CNN} \\
\hline
Conv 2D ($5 \times 5$, 1, $64$) ReLU \\
\hline
Conv 2D ($5 \times 5$, 1, $32$) ReLU \\
\hline
Max-pooling ($2 \times 2$, 2) \\
\hline
Dense (1024) ReLU \\
\hline
Dense (10) \\
\hline
\end{tabular}
\quad
\begin{tabular}{|c|}
\hline
\textbf{CIFAR-10 CNN} \\
\hline
Conv 2D ($3 \times 3$, 1, $32$) ReLU \\
\hline
Conv 2D ($3 \times 3$, 1, $32$) ReLU \\
\hline
Max-pooling ($2 \times 2$, 2) \\
\hline
Conv 2D ($3 \times 3$, 1, $64$) ReLU \\
\hline
Conv 2D ($3 \times 3$, 1, $64$) ReLU \\
\hline
Max-pooling ($2 \times 2$, 2) \\
\hline
Dense (256) ReLU \\
\hline
Dense (10) \\
\hline
\end{tabular}
\end{center}

\subsection{VAE-GANs} \label{section:settings vae-gans}

We used a common encoder structure for MNIST, FMNIST and CIFAR-10. We set the latent dimension $zdim = 10$ for MNIST and FMNIST and $zdim = 128$ for CIFAR-10. In contrast to \citet{Larsen2016}, for the reconstruction loss, we use the $\ell_1$ distance, not the discriminator's features. The architectures for VAE-GANs are given in the tables below. We reuse the notation for classification networks. Additionally, BN indicates batch normalization and for transposed convolution layers, we write \enquote{Deconv 2D (\textit{filter size}, \textit{stride}, \textit{number of filters})}.

\begin{center}
\begin{tabular}{|c|}
\hline
\textbf{Encoder} \\
\hline
Conv 2D ($4 \times 4$, 2, $64$) BN ReLU \\
\hline
Conv 2D ($4 \times 4$, 2, $128$) BN ReLU \\
\hline
Conv 2D ($4 \times 4$, 2, $256$) BN ReLU \\
\hline
Dense ($zdim * 2$) BN \\
\hline
\end{tabular}
\end{center}

\newpage

\textbf{MNIST / FMNIST.} We used the decoder and discriminator structure given in \citet{Larsen2016}. We trained the network with Adam with learning rate $\alpha = 0.0005$, $\beta_1 = 0.5$, $\beta_2 = 0.999$ and 1 discriminator update per encoder and decoder update for 30 epochs for MNIST and 60 epochs for FMNIST.

\begin{center}
\begin{tabular}{|c|}
\hline
\textbf{Decoder} \\
\hline
Dense (1024) BN ReLU \\
\hline
Deconv 2D ($4 \times 4$, 2, $256$) BN ReLU \\
\hline
Deconv 2D ($4 \times 4$, 1, $128$) BN ReLU \\
\hline
Deconv 2D ($4 \times 4$, 2, $32$) BN ReLU \\
\hline
Deconv 2D ($4 \times 4$, 2, $1$) TanH \\
\hline
\end{tabular}
\quad
\begin{tabular}{|c|}
\hline
\textbf{Discriminator} \\
\hline
Conv 2D ($4 \times 4$, 2, $32$) ReLU \\
\hline
Conv 2D ($4 \times 4$, 2, $128$) BN ReLU \\
\hline
Conv 2D ($4 \times 4$, 2, $256$) BN ReLU \\
\hline
Conv 2D ($4 \times 4$, 2, $256$) BN ReLU \\
\hline
Dense (512) BN ReLU \\
\hline
Dense (1) Sigmoid \\
\hline
\end{tabular}
\end{center}

\textbf{CIFAR-10.} We used the decoder and discriminator structure given in \citet{Miyato2018}. In the discriminator, we used spectral normalization (SN) with leaky ReLU (lReLU) activation functions with slopes set to 0.1. We trained the network with Adam learning rate $\alpha = 0.0002$, $\beta_1 = 0$, $\beta_2 = 0.9$ and 5 discriminator updates per encoder and decoder update for 150 epochs.

\begin{center}
\begin{tabular}{|c|}
\hline
\textbf{Decoder} \\
\hline
Dense (8192) \\
\hline
Deconv 2D ($4 \times 4$, 2, $256$) BN ReLU \\
\hline
Deconv 2D ($4 \times 4$, 2, $128$) BN ReLU \\
\hline
Deconv 2D ($4 \times 4$, 2, $64$) BN ReLU \\
\hline
Conv 2D ($3 \times 3$, 1, $3$) TanH \\
\hline
\end{tabular}
\quad
\begin{tabular}{|c|}
\hline
\textbf{Discriminator} \\
\hline
Conv 2D ($3 \times 3$, 1, $64$) SN lReLU \\
\hline
Conv 2D ($4 \times 4$, 2, $64$) SN lReLU \\
\hline
Conv 2D ($3 \times 3$, 1, $128$) SN lReLU \\
\hline
Conv 2D ($4 \times 4$, 2, $128$) SN lReLU \\
\hline
Conv 2D ($3 \times 3$, 1, $256$) SN lReLU \\
\hline
Conv 2D ($4 \times 4$, 2, $256$) SN lReLU \\
\hline
Conv 2D ($3 \times 3$, 1, $512$) SN lReLU \\
\hline
Dense (1) Sigmoid \\
\hline
\end{tabular}
\end{center}

\section{Adversarial Attack and Adversarial Training Settings} \label{section:adv attack and training}

All adversarial attacks in this paper were optimized to maximize the cross entropy loss or the CW surrogate objective with 40 iterations of PGD. Following previous works \citet{Tsipras2018} and \citet{Stutz2018}, during adversarial training, we trained on adversarial images only. That is, we did not mix natural and adversarial images. We describe the adversarial training procedure and attack settings used in each section.

\subsection{Section \ref{section:visual sub1}.} \label{section:2.1 settings}

For MNIST and FMNIST, we trained DNNs against $\ell_2$-bounded attacks with $\epsilon \in \{ 1, 2, 4 \}$ and $\ell_\infty$-bounded attacks with $\epsilon \in \{ 0.05, 0.1, 0.2 \}$. For CIFAR-10, we trained DNNs against $\ell_2$-bounded attacks with $\epsilon \in \{ 0.1, 0.4, 1.6 \}$ and $\ell_\infty$-bounded attacks with $\epsilon \in \{ 0.005, 0.01, 0.02 \}$. The adversarial examples used for analysis are $\ell_2$-bounded attacks with $\epsilon = 12$ which maximize the cross entropy loss.

\subsection{Section \ref{section:visual sub2}.} \label{section:2.2 settings}

We trained the networks against $\ell_2$-bounded attacks with $\epsilon = 0.25$ (weak) and $\epsilon = 0.5$ (strong). We visualized $\ell_2$-bounded attacks with $\epsilon = 1$ which maximize the cross entropy loss.

\subsection{Sections \ref{section:quantitative sub2} and \ref{section:quantitative sub3}.} \label{section:3 settings}

We trained DNNs against $\ell_2$ or $\ell_\infty$-bounded attacks of varying $\epsilon$. For $\ell_2$-bounded attacks, we linearly interpolated $\epsilon$ between $0$ and $1.6$ with step size $0.08$. For $\ell_\infty$-bounded attacks, we linearly interpolated $\epsilon$ between $0$ and $0.4$ with step size $0.002$.

\end{document}